\documentclass{llncs}
\usepackage{makeidx}  % allows for indexgeneration
\usepackage[numbers]{natbib}
\begin{document}
%
% \frontmatter          % for the preliminaries
%
\pagestyle{headings}  % switches on printing of running heads
% \addtocmark{Hamiltonian Mechanics} % additional mark in the TOC
%
%
%
\mainmatter              % start of the contributions
\title{How can deep learning advance computational modeling of sensory information processing?}
%
% \titlerunning{}  % abbreviated title (for running head)
\author{Jessica~A.F.~Thompson\inst{1,2} \and Yoshua~Bengio\inst{2} \and Elia~Formisano\inst{3} \and Marc~Sch\"{o}nwiesner\inst{1,4}}
\authorrunning{Jessica Thompson et al.} % abbreviated author list (for running head)
% \tocauthor{Ivar Ekeland, Roger Temam, Jeffrey Dean, David Grove,
% Craig Chambers, Kim B. Bruce, and Elisa Bertino}
%
\institute{International Laboratory for Brain, Music and Sound, University of Montreal, Canada\\
\and
Montreal Institute for Learning Algorithms, Montreal, Canada\\
\and Department of Cognitive Neuroscience, Maastricht University, Maastricht, Netherlands
\and
Institute for Biology, Leipzig University, Leipzig, Germany\\
}

\maketitle              % typeset the title of the contribution

\begin{abstract}
Deep learning, computational neuroscience, and cognitive science have overlapping goals related to understanding intelligence such that perception and behaviour can be simulated in computational systems. In neuroimaging, machine learning methods have been used to test computational models of sensory information processing. Recently, these model comparison techniques have been used to evaluate deep neural networks (DNNs) as models of sensory information processing. However, the interpretation of such model evaluations is muddied by imprecise statistical conclusions. Here, we make explicit the types of conclusions that can be drawn from these existing model comparison techniques and how these conclusions change when the model in question is a DNN. We discuss how DNNs are amenable to new model comparison techniques that allow for stronger conclusions to be made about the computational mechanisms underlying sensory information processing.

\end{abstract}
\section{The goal of information-based neuroimage analysis}
In the last 15 years, information-based approaches have become a popular technique for analyzing neuroimaging data. Typically this involves evaluating a model of human perception or cognition on its ability to predict some measure of brain activity. In studies of sensory information processing, the model usually consists of some signal processing applied to the stimuli. Recently, researchers have begun using deep neural networks (DNN)s, trained on large datasets, as models to be evaluated in this information based-approach.
%Some of these networks are very intelligent and can achieve near or even superior to human performance on tasks like visual object recognition (the winners of the ILSVRC 2015 classification task achieved an error rate of 3.57\% with a network with 152 layers \citep{He2015} \footnote{Anrej Karpathy estimates his own performance on the task to be 5.1\%. See http://karpathy.github.io/2014/09/02/what-i-learned-from-competing-against-a-convnet-on-imagenet/}).
%But what if we had a model that perfectly replicated human performance on a set of tasks. What would that gain us in this approach?
How might this approach lead to a better understanding of sensory information processing?
%\todo{Do these questions map to the sections of the following text? That would be a neat 'intro'}
What types of conclusions can be drawn from information-based approaches and are those conclusions dependent on the type of model being evaluated? To answer these questions, we must first consider the nature of the understanding that we wish to extract from these analyses.
%\todo{You are at the stage of collecting and sorting ideas and structuring the text, but I will nevertheless also comment on the general style. My impression of the first few paragraphes is that you can increase clarity by tightening up the text. It's a bit wordy. Avoid too much 'us'. Are the footnotes really relevant?}

Several authors have recently discussed ``what constitutes an explanation?''
A common theme is to invoke \citeauthor{Marr1982}'s levels of analysis \citep{Poeppel2012, Love2015, Yamins2016, Jonas2016,Buzsaki2006,Pratte2016,Varoquaux2014}. The first level, \textit{computational theory}, is concerned with the `what' and `why'. What is the system doing and why is it doing it? The next level is \textit{representation and algorithm}, which is concerned with `how' the computational theory is carried out. What are the input and output representations and what is the algorithm that transforms one into the other? The last level is the \textit{hardware implementation}, which is concerned with the physical realization of the representation and algorithm. Marr argues that these different layers can be studied separately but ultimately should be consistent with one another for a system to be understood \citep{Marr1982}. \Citeauthor{Poeppel2012} argues that there is currently a granularity mismatch between the units of neurobiology and the units of cognition and perception. He suggests that we should focus on the representation and algorithm level to help bridge this mismatch \citeyearpar{Poeppel2012}. \Citet{Love2015} has also suggested that the \textit{inside-out} approach, starting with the representation and algorithm level and working upward to the computational level and downward to the implementation level, is best way to achieve compatible understanding at all three layers. Evaluating computational models on their correspondence to neuroimaging data can be viewed as a way to bridge the representational and implementation levels of understanding \citep{Love2015}.%\todo{I think this paragraph in particular can be shortened a lot by distilling the ideas a bit more, right now it sounds like an essay on Poeppel's paper.}

%\todo{Good example of a sentence in conversational style (I can see you saying this), I suggest striving to be more concise and perhaps a bit more formal.}
\Citet{Marr1978} define a representation as ``a formal system for making explicit certain entities or types of information, together with a specification of how the system does this.'' For example, the Arabic numeral 37 is a description of the number 37 that makes explicit its decomposition into powers of ten. A binary representation of the same number would make explicit its decomposition into powers of two. A representation will often be a useful abstraction. For example, we can represent strands of DNA as sequences of nucleotides, represented by the letters A, T, C and G. Similarly, the brain is likely to use various representations of sensory information at different stages to facilitate computation. %In information-based analyses of neuroimaging, we assume a) that these neural representations are activity based \todo{I don't understand what you mean here. A type of neural code? What other option would there be?} and b) that our measurements reveal something about neural representations.
%\todo{maybe mention limitations of fMRI e.g. \citep{Logothetis2008} haemodynamic response sensitive to the size of activated population and therefore less able to reveal sparse representations.}\todo{Amir Shmuel at the MNI has nice data about differenciating the different hypotheses of how voxel activity samples the underlying neural responses. Can this sampling be part of the model?}

\section{What can be learned from information-based analyses?}
%\todo{Be careful not to overuse the rethoric questions.}
In studying sensory information processing, we are concerned with how brains ``represent and recognize statistical invariances in the environment that guide its actions, ultimately ensuring our survival in a world that is in a continuous state of flux''\citep{VanGerven2016}. Several approaches have been used to explore the information content of neural responses. We will briefly review a few of these approaches and clarify some common misconceptions about their use.
% \subsection{MVPA, Encoding and RSA}

Multi-variate pattern analysis (MVPA) uses machine learning algorithms to classify or predict experimental conditions or stimulus properties from multi-variate patterns of activity \citep{Norman2006}. This type of analysis performs an omnibus test of whether a pattern of activity contains any information about a specific stimulus property \citep{Haynes2015}. However, this method is problematic for the comparison of computational models due to the geometric, spatial and representational ambiguity inherent to the approach \citep{Naselaris2011}.
%``decoding performance does not directly indicate the amount of variance in activity that is attributable to a given feature. A feature may be perfectly decoded from population activity even though it is responsible for little variance in activity. \dots Therefore, comparing classification performance across different kinds of feature is an ‘apples-to-oranges’ comparison that is likely to mislead.'' \citep{Naselaris2011}
% MVPA Beginning of information-based neuroimage analysis. Omnibus test of non-zero mutual information. Problem: MVPA can only provide partial functional description of region of interest \citep{Formisano2008}

An alternative is to build encoding models wherein activity in each measurement channel is modeled as a combination of features representing the stimuli. When fitting an encoding model to data, the approach has also been called reverse correlation and situated in the context of system identification \citep{Wu2006}. Because this approach typically models each channel separately, it cannot capture information in spatially distributed patterns of activity and suffers from a multiple comparisons problem. However, An encoding model can sometimes be inverted to get at multivariate information while still modeling channels independently \citep{Nishimoto2011}.
% add \citep{Santoro????}

A complementary approach, representational similarity analysis (RSA), abstracts away from measurement channels and instead works on activity dissimilarity matrices (DSMs) \cite{Kriegeskorte2008}. This allows for easy comparison across modalities, species, and models while also including multivariate information. No model fit to the data is needed. RSA is ideal for high-dimensional data because the number of experimental conditions is usually less than the number of channels \citep{Kriegeskorte2009,Kriegeskorte2013}. An extension of RSA, pattern-component modeling (PCM), operates on the pattern components of the DSMs, further enabling comparisons between different regions and individuals \citep{Diedrichsen2011}.

\citet{Diedrichsen2016} present a common framework for understanding encoding, representational-similarity and pattern-component analysis. They note that these methods all calculate the statistical second moment of the activity profile distribution, which contains all information about any differences in mean and variance across experimental conditions (under the condition of i.i.d. noise). Their definition of representation within this framework is that a represented variable can be linearly decoded from a down-stream area. They suggest that focusing on linear encoding as a starting point for evaluating any representation model.

We agree that linear encoding is a good starting point but would like to clarify the types of conclusions that can be drawn from these methods and propose a different working definition of representation. From a representation learning perspective, a representation defines the feature space in which some data live. A series of representational transformations can be viewed as \textit{disentangling} the underlying factors of variation to make some specific information more easily accessible. However, a representation is not defined by a single piece of information but rather by the shape of the manifold on which the data lie within the representational space \citep{Bengio2013}. Therefore, we propose the following qualification to the types of conclusions that are drawn from encoding and RSA.

Proponents of encoding often make claims of the type ``X stimulus property is encoded in brain area Y'' because they were able to predict responses in Y from X representation of the stimulus (better than some other model Z). The desired interpretation of this type of result is that the computational role of region Y is to calculate or represent X. Even if model X captured 100\% of the variance in Y, this result is only a partial description of the information content of the activity measured in Y. For example, mel-frequency cepstral coefficients (MFCCs) are a commonly used acoustic feature for various machine hearing tasks. Linear classifiers can be trained on MFCCs to predict vowels \citep{Li2005}, musical instrument \citep{Herrera-Boyer2003}, and musical genre \citep{Tzanetakis2002}. But MFCCs are not a representation of vowels, instruments, or genre. From a computational perspective, the difference between MFCCs and some representation of genre matters. Therefore, if MFCCs were able to predict activity in some brain region, one cannot separate whether the brain region is using a representation similar to MFCCs or if it is encoding some other property that can be linearly decoded from MFCCs.

Similarly in RSA, authors sometimes make claims such as “brain area X uses a representation similar to that of model Y” because their DSMs were highly correlated. Such claims are imprecise because there are infinitely many stimulus representations that will result in the same DSM. Any rotation or translation of the data within the original feature space will lead to an identical DSM. \citet{Dutilleul2000} have shown that if bivariate data are generated by a multivariate Gaussian model and have a spherical variance-covariance structure, then the correlation of the data will be quadratically related to the the correlation of their DSMs. However, they performed simulations to show that this does not hold for other types of models (mixtures of Gaussians, exponential, etc.) and demonstrated with empirical results that DSMs may show significant correlation even when the correlation of the data vectors is zero. Therefore, it is important to be clear that RSA reveals the similarity of representational geometries---the relative distinction between stimuli according to a particular distance metric---and says very little about the form of the specific neural representation, which have been abstracted away from in RSA.

We can distinguish between the information content of a representation and its representational \textit{form}. For example, the numerals 37 and XXXVII are equally identifiable as the number 37, but the form of the representation is different. \Citet{Diedrichsen2016} point out that ``knowing the content and format of representations provides strong constraints for computational models of brain information processing''\citep{Diedrichsen2016}. Indeed, authors often motivate and interpret encoding or RSA style analyses with questions about representational form. For example, \citet{Turner2016} writes, ``results using [an encoding] approach reveal that specific features of experience often have widely distributed spatial representations in the brain''. These types of claims can be misleading because accurate linear predictions do not reveal the form of neural representations. Several candidate stimulus representations may obtain identical encoding performance in a given region. Conversely, a candidate stimulus representation may predict activity equally well in large swaths of brain regions. Accurate linear predictions reveal that a neural representation lives in some space of all possible linear transformations of the model representation. This allows us to conclude that there is some shared information between the model and the brain but does not allow for conclusions about the form of neural representations.

MVPA, encoding and RSA make partial descriptions of the information content of neural representations. Therefore, these are complementary approaches, as mentioned in \citep{Diedrichsen2016}. Unambiguous description of the information content of a neural representation would mean finding representations that can simultaneously fully explain neural activity and be fully explained by the same neural activity such that no variance in either domains is unaccounted for. We suggest that new methods are necessary to describe the form of neural representations and that DNNs may be especially amenable to new model evaluation techniques that enable such description.

\section{What can be learned from DNNs?}
% \subsection{DNNs as models of sensory information processing}
Recently, DNNs have been used as representational models in encoding and RSA analyses of auditory \citep{Guclu2016} and visual \citep{Yamins2013, Agrawal2014, Cadieu2014, Khaligh-Razavi2014, Guclu2015} responses. The use of DNNs as models of sensory processing allows for hypotheses to be embedded in the design of the network rather than in explicitly selected stimulus representations. In this approach, typically each layer of a trained network is treated as a representational model and researchers look for correspondence between DNN layers and brain regions \citep{Kriegeskorte2015}. \Citet{Yamins2016} point out that using DNNs as models of sensory information processing is in line with Marr's levels of understanding. Unlike other unlearned representational models, goal-driven or task-optimized models are functionally constrained to be able to perform some task. This approach seems to epitomize \citeauthor{Love2015}'s inside-out approach in that hypotheses about neural representations are constrained both by computational theory (what the system does) and implementation (how well does it align with neural measurements?).
However, the usefulness of DNNs as models of neural information processing has been questioned for several reasons, such as their biological implausibility (which has been well addressed in \citep{Kriegeskorte2015, Marblestone2016, Yamins2016}) and the fact that they are difficult to interpret. Are linear models really preferable to DNNs because they are more interpretable? \Citet{Lipton2016} suggests that such claims about interpretability need to be qualified because interpretability is not a monolithic concept, but rather it can be divided into specific desirable attributes. They consider several properties of interpretable models related to transparency (i.e. how does the model work?) and post-hoc interpretations (i.e. what can the model tell me?). They conclude that linear models are not strictly more interpretable than DNNs. For example, ``large linear models may be just as opaque as deep neural networks. However, for post-hoc interpretability, deep neural networks exhibit a clear advantage, learning rich representations that could be visualized, verbalized, or used for clustering'' \citep{Lipton2016}.

Particularly in sensory neuroscience, researchers may ask ``do DNNs learn meaningful representations?'' as a prerequisite for their usefulness. To some extent, this reflects the belief that in order for DNN to be useful, it should recover what is already known about neural sensory information processing. Only then might the DNN help to fill in the gaps of our knowledge. It also seems to reflect an expectation that DNNs learn features that can be understood, meaning attributable to some property of the physical world or higher-level cognitive concept. \Citet{Kriegeskorte2015} points out that ``a verbal functional interpretation of a unit, e.g., as an eye or a face detector, may help our intuitive understanding and capture something important. However, such verbal interpretations may overstate the degree of categoricality and localization, and understate the statistical and distributed nature of these representations.'' The brain's capacity to map from sensory input to high-level concepts will necessarily involve intermediate representations that are not directly related to properties of the physical world of objects of cognition. It is an advantage of DNNs that they also demonstrate this complexity.

The ``deep'' aspect of deep neural networks also presents a challenge in that each layer of a DNN can be viewed as a distinct representational model. Previous work has mapped individual layers to different brain areas by comparing encoding or RSA performance across layers and regions, but this still treats layers as independent models. A good model will make accurate predictions across the entire sensory processing pathway, not just at individual layers/regions, so ideally entire networks would be evaluated rather than individual layers. The solution to this challenge may also enable characterizations of representational form. \Citet{Peters2016} advocate for the ``direct integration'' of models and brain activity by projecting them both into a common space, allowing for evaluations within and across processing stages \citep{Peters2012}. DNNs may be especially amenable to such an approach because they can be designed to have correspondence with a particular set of regions of interest. Additionally, graph theoretical methods have been developed allowing for definitions of brain graphs from functional, structural and or diffusion weighted images \citep{Bullmore2009}. Such brain graphs could be used as templates for DNNs, providing a natural common space in which model and brain could be compared directly. Such an approach could allow for the characterization of representational form as well as content.

\section{Conclusion}
The goal of information-based neuroimage analysis is to describe human intelligence at the representation and algorithm level of understanding. These methods can be seen as part of an inside-out approach wherein models are constrained by their performance (computational theory) and their correspondence to brain activity (implementation). MVPA, encoding, and RSA yield complementary characterizations of the information content of neural representations. However, these methods cannot easily characterize representational form. DNNs are useful as models of sensory information processing because they can be designed to demonstrate certain behaviours or structures of biological brains while allowing unknown characteristics to be learned from data. This allows DNNs to discover candidate representations in a data-driven manner that are constrained by the behaviour of the network and their similarity (in form and content) to measured neural representations. By constructing intelligent artificial networks that demonstrate increasing biological realism and also modeling our neural measurements as networks, we can begin to transfer knowledge about artificial networks to knowledge about biological networks and vice versa.

% A satisfactory explanation of the neural basis of perception and cognition will consist of understanding at (at least) the three levels outlined by Marr. Ultimately these levels of understanding should be consistent with one another, but importantly, a description at one level does not imply a specific description at another level.

% Artificial neural networks can be useful because
% better performing (at understanding the world and predicting brain activity) than existing models
% -can discover representations that we would not arrive it via hypothesis driven approach
% -can model entire functional pathways
% -they can learn - can observe how representations change during learning,
% -can be constrained/designed to demonstrate certain characteristics of biological brains while allowing unknown characteristics to be learned
% -allow for model evaluation

\bibliographystyle{authordate1}
\bibliography{mlini2016}

\end{document}